\documentclass[runningheads]{llncs}

\usepackage{iftex}
\ifpdftex\usepackage[T1]{fontenc}\fi

\usepackage{graphicx}
\usepackage{amsmath,amssymb}
\usepackage{booktabs}
\usepackage{array}
\usepackage{color}
\usepackage{bm}
\usepackage{multirow}
\usepackage{makecell}
\usepackage{placeins}
\usepackage{bbding}
\usepackage{url}
\AtBeginDocument{\def\doi#1{\url{https://doi.org/#1}}}
\makeatletter
\renewcommand\paragraph{\@startsection{paragraph}{4}{\z@}%
        {-4\p@ \@plus -2\p@ \@minus -2\p@}%
        {-0.5em \@plus -0.22em \@minus -0.1em}%
        {\normalfont\normalsize\itshape}}
\makeatother
\newcommand{\Y}{\mathbf{Y}}
\newcommand{\A}{\mathbf{A}}
\newcommand{\X}{\mathbf{X}}
\newcommand{\Snoise}{\mathbf{S}}
\newcommand{\Nnoise}{\mathbf{N}}
\newcommand{\R}{\mathbf{R}}

\newcommand{\Ve}{\mathbf{V}}
\newcommand{\Ee}{\mathbf{E}}
\newcommand{\Ze}{\mathbf{Z}}
\newcommand{\Hh}{\mathbf{H}}
\newcommand{\prox}{\operatorname{prox}}

\begin{document}

\title{AXS-Net: Interpretable Deep Unfolding for Hyperspectral Image
Denoising via Spectral Basis Unmixing and Structured Noise
Refinement\thanks{This work is accepted to the 14th International Conference on Image and Graphics (ICIG2026), 2026, and partly supported by Hunan Provincial Natural Science
Foundation of China, under the Science and Technology Innovation Program of
Hunan Province (Project No.~2025JJ60883).}}
\titlerunning{AXS-Net: Interpretable Unfolding for HSI Denoising}

\author{Ziyi Guan \and
Jianping Zhang\textsuperscript{(\Envelope)} \and
Zheng Yang}
\authorrunning{Z. Guan et al.}
\institute{Xiangtan University, Xiangtan, Hunan, China\\
\email{ziyiguanxtu@163.com},
\email{jpzhang@xtu.edu.cn},
\email{202421511256@smail.xtu.edu.cn}}

\maketitle

% The abstract should briefly summarize the paper in 150--250 words.
\begin{abstract}
Hyperspectral images (HSIs) are often degraded by mixed noise, including band-dependent Gaussian perturbations and structured artifacts such as stripes, dead-lines, and impulse noise. Most deep denoisers regress the clean image directly, entangling signal and structured noise. We instead model HSI denoising as $\Y=\A\X+\Snoise+\Nnoise$, where $\A\X$ is a low-rank spectral-subspace (unmixing) reconstruction, $\Snoise$ is structured sparse noise and $\Nnoise$ is residual Gaussian noise. The resulting regularized optimization problem is unrolled into AXS-Net, a $K$-stage alternating proximal-point framework. Each stage combines an analytic spectral-basis gradient step, an SSX-Block proximal operator for abundance coefficients, and an SBlock proximal operator for the structured residual with column-consistent and sparse priors. This optimization correspondence exposes interpretable endmembers, abundance maps, and structured-noise estimates. Across ICVL, CAVE, and Harvard datasets and five noise configurations, the proposed AXS-Net achieves strong in-domain accuracy and competitive zero-shot transfer, with consistent gains across all five noise regimes on ICVL and Harvard. The recovered structured-noise closely follows the synthetic reference, and the recovered spectral basis is smooth and band-ordered rather than an arbitrary set of latent channels.

\end{abstract}

\section{Introduction}\label{sec:intro}
Hyperspectral images (HSIs) capture a dense spectrum at every spatial location and support remote sensing, material identification, and computational imaging. During acquisition, however, they are routinely degraded by mixed noise, consisting of band-dependent Gaussian perturbations and structured corruptions such as stripes, dead-lines, and impulse noise. These artifacts not only reduce visual quality but also distort per-pixel spectral characteristics, making mixed-noise removal a fundamental preprocessing problem for downstream tasks \cite{lrmr}.

Classical model-based denoising methods restore HSIs with explicit hand-crafted priors such as low-rankness or total-variation terms~\cite{lrmr,lrtv} and non-local self-similarity~\cite{bm4d}, optionally combined with global spectral low-rankness~\cite{ngmeet}. They are interpretable and can separate structured terms, but fixed priors and iterative solvers limit accuracy and efficiency. Learning-based methods learn stronger spatial--spectral representations~\cite{hsdt,ssumamba} by using sparse-coding and model-aided designs~\cite{t3sc,macnet}, transformers~\cite{hsdt}, and channel/spectral attention~\cite{senet}, yet most regress a single clean image, entangling the signal with structured noise and offering no explicit handle on stripes or impulses. Deep unfolding embeds iterative solver steps of the optimization problem into trainable networks~\cite{istanet}, combining model structure with learned capacity. However, existing unfolded denoisers still optimize a single clean-image variable, leaving the structured corruption implicit. Classical low-rank-plus-sparse formulations usually rely on fixed sparse outlier priors and do not provide a learned, separately inspectable structured-noise branch coupled with an unmixing-style clean component.

In this work, we unroll the alternating minimization of a hyperspectral low-rank-plus-sparse decomposition into \textbf{AXS-Net}, whose stages comprise an analytic spectral-basis update, an SSX-Block for abundance estimation, and an SBlock for structured-residual refinement, rather than producing only a denoised image from observation $\Y$. Our contributions are summarized as follows.
\begin{itemize}
  \item %\textbf{Interpretable spectral-subspace decomposition.} 
  We formulate HSI mixed-noise removal as $\Y=\A\X+\Snoise+\Nnoise$ and unroll its
  alternating proximal solution into an interpretable $K$-stage network, separating an unmixing-style clean component $\A\X$ from a dedicated
  structured-noise estimate $\Snoise$ instead of regressing a single clean image.
  \item %\textbf{Two learned proximal operators.} 
  We design the \emph{SSX-Block}, which integrates window-based spatial attention with cross-spectral attention to refine abundances, and the \emph{SBlock}, which fuses directional residual features with parameter-free column-consistency and sparsity priors plus adaptive thresholding to separate structured noise.
  \item %\textbf{Strong restoration with examined interpretability.} 
  On the ICVL, CAVE, and Harvard datasets with Gaussian, stripe, dead-line, impulse, and mixed noise, our AXS-Net achieves top performance on in-domain ICVL and generalizes effectively to the unseen CAVE and Harvard datasets, with the largest margins on Harvard. At the same time, its recovered structured noise closely follows the synthetic ground truth, and the stage-wise decomposition remains spectrally organized and interpretable throughout the unrolled iterations.
\end{itemize}

\section{Methodology}
\label{sec:method}

\subsection{Problem Formulation}
\paragraph{Observation model.}
Let $\Y_{o}\in\mathbb{R}^{s\times h\times w}$ denote the observed HSI with $s$ spectral bands and spatial size $h\times w$. Since the spectra of a natural scene are highly correlated across bands, the clean signal can be represented by a small number of shared spectral signatures. Following the linear spectral mixing model~\cite{unmix}, we assume that the underlying clean HSI lies in a low-dimensional spectral subspace, i.e., each clean pixel can be approximated by a non-negative combination of $r\ll s$ representative spectra, or endmembers. Whenever matrix factorization is involved, we flatten the tensor $\Y_{o}$ spatially into $\Y\in\mathbb{R}^{s\times n}$ with $n=hw$, and model the observation as
\begin{equation}
  \Y = \A\X + \Snoise + \Nnoise,
  \label{eq:model}
\end{equation}
where $\A\in\mathbb{R}^{s\times r}$ is the spectral basis, or endmember matrix whose columns are the shared spectra, $\X\in\mathbb{R}^{r\times n}$ is the abundance matrix collecting the pixel mixing coefficients, $\Snoise$ denotes structured sparse noise that captures corruptions such as stripes, dead-lines, and impulses, and $\Nnoise$ denotes residual zero-mean Gaussian noise. This three-component decomposition~\cite{lrmr} assigns a distinct role to each term: the low-rank product $\A\X$ reconstructs the clean image in an unmixing style and is taken as the denoised output, while $\Snoise$ is \emph{explicitly retained} rather than being absorbed into a single regression output, so that structured corruptions are exposed as an interpretable and separately inspectable estimate.

\paragraph{Constrained optimization.}
Given this model, we recover $\A$, $\X$, and $\Snoise$ by minimizing the residual of reconstruction under suitable regularization priors:
\begin{equation}
\begin{aligned}
  \min_{\A,\X,\Snoise}\quad
  & \frac{1}{2}\left\|\Y-\A\X-\Snoise\right\|_F^2
    + \lambda_X \mathcal{R}_X(\X)
    + \lambda_S \mathcal{R}_S(\Snoise),\\
  \text{s.t.}\quad
  & \A \ge 0,\qquad
    \left\|\A_{:,j}\right\|_2 = 1,\quad j=1,\ldots,r,\qquad
    \X \ge 0.
\end{aligned}
  \label{eq:objective}
\end{equation}
The quadratic term $\tfrac12\left\|\Y-\A\X-\Snoise\right\|_F^2$ serves as the data-fidelity term induced by Gaussian noise $\Nnoise$, while the priors $\mathcal{R}_X$ and $\mathcal{R}_S$, weighted by $\lambda_X$ and $\lambda_S$, regularize the abundances and the structured noise, respectively. The
constraints make the factorization physically meaningful. Specifically, $\A\ge0$ together with the unit-$\ell_2$ column normalization $\left\|\A_{:,j}\right\|_2=1$ forces each endmember to be a non-negative and scale-fixed spectral basis, as in non-negative matrix
factorization~\cite{nmf}. The constraint $\X\ge0$ keeps the abundances non-negative, consistent with their interpretation as mixing proportions.

\paragraph{Learned proximal operators and initialization.}
The task-specific regularizers $\mathcal{R}_X$ and $\mathcal{R}_S$ are difficult to specify in closed form: abundance maps exhibit complex spatial--spectral structures, and structured noise may contain multiple corruption types with unknown statistics. We therefore keep the proximal-gradient \emph{structure} of the solver but replace the two proximal operators $\prox_{\lambda_X\mathcal{R}_X}$ and $\prox_{\lambda_S\mathcal{R}_S}$ with learnable modules trained end-to-end (Sec.~\ref{sec:framework}), where the regularization weights $\lambda_X,\lambda_S$ are absorbed into these modules. The proximal step
$\prox_{\lambda\mathcal{R}}(\mathbf{v})=\arg\min_{\mathbf{u}}\tfrac12\|\mathbf{u}-\mathbf{v}\|_F^2+\lambda\mathcal{R}(\mathbf{u})$ can be interpreted as a denoising or MAP subproblem, whose solution is refined by the
prior $\mathcal{R}$. Because the proximal mappings $\prox_{\lambda\mathcal{R}}(\mathbf{v})$ associated with the task-specific priors $\mathcal{R}_X$ and $\mathcal{R}_S$ do not admit simple closed-form expressions, we adopt the deep unfolding approach~\cite{istanet} and represent these proximal
operators using neural networks (SSX-Block and SBlock). At the same time, we retain the analytic forms for the
decomposition components, including the $\A$-step and the residual $\R^{k}$. To start the alternating scheme from a subspace-informed point, we construct $\A_0$ directly from the data. Specifically, we first mean-center $\Y$ across the spatial dimension, then perform an SVD to extract the top-$r$ left singular vectors and use these leading spectral components as an initial basis. This basis is subsequently mapped onto the feasible set via a non-negativity projection followed by column-wise $\ell_2$ normalization to the unit norm, resulting in $\A^0$. The remaining variables are initialized as $\X^0=\operatorname{ReLU}\!\big({\A^0}^{\!\top}\Y\big)$ and $\Snoise^0=\mathbf{0}$. These quantities are progressively refined by the unrolled stages described below.

\begin{figure}[ht]\centering
\includegraphics[width=0.92\textwidth]{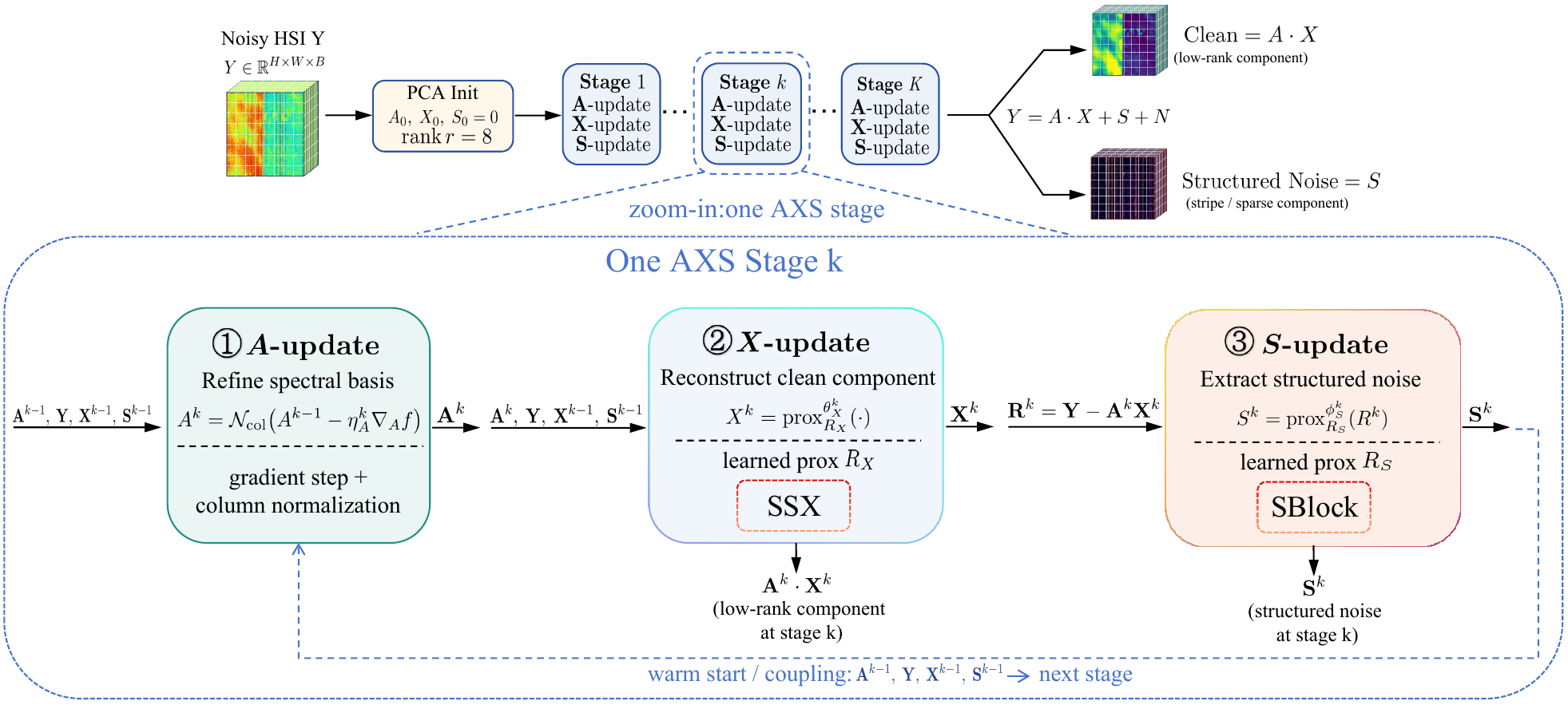}
\caption{Overview of AXS-Net. From a PCA (mean-centered SVD) subspace initialization, $K{=}6$ unrolled stages alternately update the spectral basis $\A$, abundances $\X$ (SSX-Block) and structured noise $\Snoise$ (SBlock), extracting the clean component $\A\X$ and $\Snoise$ from $\Y$.}
\label{fig:framework}
\end{figure}

\subsection{Overall Unfolding Framework}
\label{sec:framework}
\paragraph{From optimization to a fixed-depth network.}
Problem~\eqref{eq:objective} is a non-convex constrained optimization task for two main reasons: first, the data-fidelity term couples $\A$ and $\X$ through the bilinear product $\A\X$, which makes the objective jointly
non-convex in $(\A,\X)$, although it remains convex in each variable when the other is fixed; second, the unit-$\ell_2$ equality constraint $\left\|\A_{:,j}\right\|_2=1$ confines $\A$ to a non-convex (spherical) feasible set. A natural strategy is therefore to use alternating minimization, updating one variable at a time while keeping the others fixed; each of the resulting subproblems is substantially simpler and can be solved with a straightforward projected-gradient or proximal step~\cite{ista}. Instead of running such an iterative algorithm to convergence, which is computationally expensive and still depends on manually designed regularizers, we \emph{unroll} a fixed number $K$ of alternating updates into a feed-forward network and learn the operators at each iteration directly from data~\cite{istanet}. We choose $K{=}6$ as a compromise between restoration performance and computational cost, since every additional stage introduces one more unrolled pass and thus extra latency. All $K$ stages share the same architectural design but have \emph{distinct} parameters for each stage, allowing them to specialize instead of repeatedly applying a single fixed update.

\paragraph{Stage update.} At stage $k$, the variables are refined in order $\A\!\rightarrow\!\X\!\rightarrow\!\Snoise$:
\begin{equation}
\begin{aligned}
  \A^{k} &= \mathcal{A}^{(k)}
    \!\left(\Y,\A^{k-1},\X^{k-1},\Snoise^{k-1}\right),\\
  \X^{k} &= \prox_{\lambda_X\mathcal{R}_X}
    \!\left(\Ve^{k},\Ee^{k},\X^{k-1},\Ze^{k}\right)
    && \text{(SSX-Block)},\\
  \Snoise^{k} &= \prox_{\lambda_S\mathcal{R}_S}
    \!\left(\R^{k},\Snoise^{k-1}\right),\quad
    \R^{k} = \Y-\A^{k}\X^{k}.
\end{aligned}
  \label{eq:stage_overview}
\end{equation}
The $\A$-update remains an analytic gradient step, whereas SSX-Block and SBlock learn the proximal updates for $\X$ and $\Snoise$, respectively (Secs.~\ref{sec:ablock}--\ref{sec:sblock}), preserving the model structure while learning priors that are difficult to specify explicitly.

\paragraph{Cross-variable coupling.}
Beyond the sequential $\A\!\rightarrow\!\X\!\rightarrow\!\Snoise$ dependency within each stage, two explicit couplings link the variables across the unrolling and are drawn as dashed arrows in Fig.~\ref{fig:framework}: (i)~the current structured-noise estimate $\Snoise^{k-1}$ is subtracted when forming the coefficient-space
residual $\Ee^{k}$, so that abundance refinement is not contaminated by structured corruptions; and (ii)~a Nesterov warm start carries momentum from
$(\X^{k-1}-\X^{k-2})$ into $\Ze^{k}$, accelerating the abundance update across stages. Together these turn the $K$ stages into a single differentiable
network trained end-to-end.

\subsection{Endmember Update: Analytic Spectral Basis Step}
\label{sec:ablock}
The endmember update $\A$ corresponds to a single unfolded projected-gradient iteration applied to the smooth data-fidelity term. In contrast to the abundance and structured-noise updates, which rely on learned proximal modules, this update employs the closed-form gradient with respect to $\A$. Starting from $\A^{k-1}$, $\X^{k-1}$, and $\Snoise^{k-1}$, we evaluate the normalized gradient
\begin{equation}
  \mathbf{G}_A^{k}
  = \frac{1}{n}\left(\A^{k-1}\X^{k-1}
      + \Snoise^{k-1} - \Y\right){\X^{k-1}}^\top,
  \label{eq:agrad}
\end{equation}
which corresponds to the gradient of the quadratic data term in Eq.~\eqref{eq:objective}. We next perform a gradient descent update and project the result back onto the feasible set:
\begin{equation}
  \A^{k}
  = \operatorname{NormCol}
    \!\left(
      \mathbf{\Pi}_{+}\!\left(\A^{k-1}-\alpha_k\mathbf{G}_A^{k}\right)
    \right),
  \label{eq:aupdate}
\end{equation}
where the projection operator $\mathbf{\Pi}_{+}$ sets all negative components to zero, and $\operatorname{NormCol}$ normalizes each column to have unit $\ell_2$ norm. Together, these operations guarantee that $\A$ satisfies both the non-negativity and unit-norm constraints. The step size is stage-dependent and learned, with the parameterization $\alpha_k=\operatorname{softplus}(\eta_k)$ enforcing positivity.

Since this subproblem is smooth, we choose to keep the A-update analytical and parameter-light instead of learning it, which preserves the interpretation of the endmember $\A$. The A-update is kept fixed during the first $10$ epochs, when the initial abundance estimates are still unstable, and activated thereafter.

\begin{figure}[!t]\centering
\includegraphics[width=0.95\textwidth]{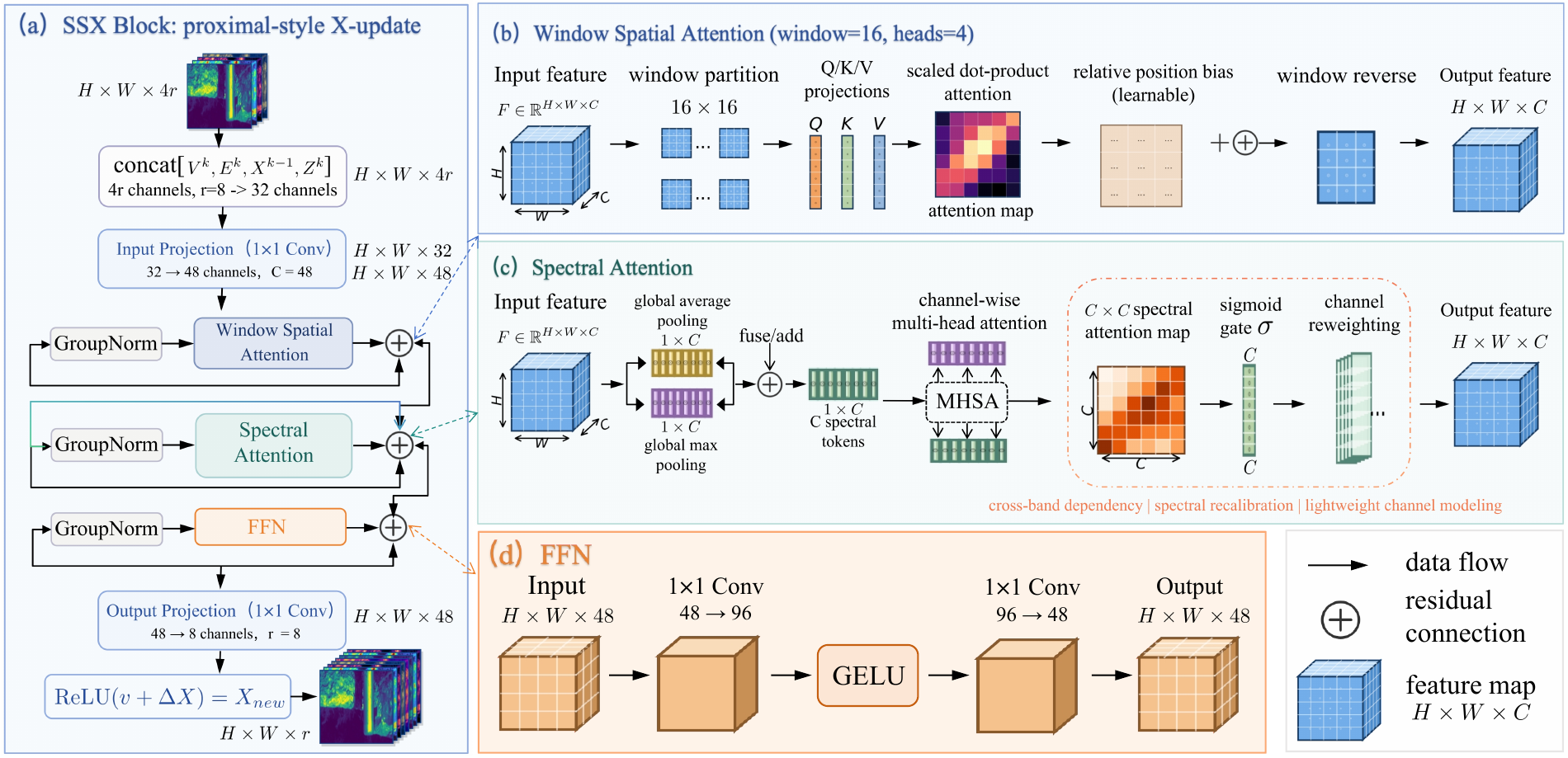}
\caption{SSX-Block for the abundance update. Inputs
$[\Ve,\Ee,\X_{\mathrm{prev}},\Ze]$ are projected to width $C$, refined by
window spatial attention, spectral attention, and an FFN, then output as
$\X^{k}=\operatorname{ReLU}(\Ve+\Delta\X)$.
}
\label{fig:ssx}
\end{figure}

\subsection{Abundance Update: Spatial--Spectral Proximal Block}% (SSX-Block)}
\label{sec:xblock}
After updating the spectral basis, we refine the abundance coefficients using a learned proximal operator applied to the gradient-corrected estimate $\Ve^{k}$. Given $\A^{k}$, $\X^{k-1}$, and $\Snoise^{k-1}$, we first compute the back-projected residual $\Ee^{k}$, a gradient-corrected iterate $\Ve^{k}$, and an extrapolated momentum feature $\Ze^{k}$ as:
\begin{align}
\Ee^{k}
  =& {\A^{k}}^\top
    \!\left(\Y-\A^{k}\X^{k-1}-\Snoise^{k-1}\right),\\
  \Ve^{k} =& \X^{k-1} + \beta_k\,\Ee^{k},\quad
\Ze^{k} = \X^{k-1} + \gamma_k\!\left(\X^{k-1}-\X^{k-2}\right),
  \label{eq:z}
\end{align}
where $\beta_k=\operatorname{softplus}(\eta_k^X)$ and $\gamma_k=\sigma(\rho_k^X)$ are learnable stage-dependent scalars. After reshaping the coefficient maps, these four tensors are concatenated into a $4r$-channel input and linearly projected to a hidden dimension $C$:
\begin{equation}
  \Hh_0=\operatorname{Conv}_{1\times1}
        \!\big(\operatorname{Concat}(\Ve^{k},\Ee^{k},\X^{k-1},\Ze^{k})\big).
  \label{eq:xinput}
\end{equation}

We realize $\prox_{\lambda_X\mathcal{R}_X}$ using a separable spatial–spectral attention module (Fig.~\ref{fig:ssx}), motivated by the fact that abundance maps exhibit both spatial self-similarity and inter-component correlations. This module consists of three pre-normalized residual sub-layers, followed by a
nonnegative residual readout:
\begin{align}
  \Hh_1 &= \Hh_0 + \operatorname{WSA}\!\big(\operatorname{GN}(\Hh_0)\big),\quad
  \Hh_2 = \Hh_1 + \operatorname{SSA}\!\big(\operatorname{GN}(\Hh_1)\big),
  \label{eq:ssa}\\
  \Hh_3 &= \Hh_2 + \operatorname{FFN}\!\big(\operatorname{GN}(\Hh_2)\big),\quad
  \X^{k} = \operatorname{ReLU}
            \!\big(\Ve^{k}+\operatorname{Conv}_{1\times1}(\Hh_3)\big),
  \label{eq:xblock}
\end{align}
where $\operatorname{GN}$ denotes GroupNorm and $\operatorname{FFN}$ is implemented as a
$1\times1$–GELU layer followed by a $1\times1$ MLP with expansion ratio $2$. Adopting a Swin-style attention mechanism~\cite{swin}, the {\em window spatial attention} (WSA) divides the feature tensor into non-overlapping $16\times16$ windows and performs $4$-head self-attention with relative-position bias in each window. This design captures local spatial relationships in the abundance maps while keeping the linear complexity in the number of windows. In analogy to channel/spectral attention methods~\cite{senet}, the {\em spectral attention} (SSA) first forms a descriptor for each channel through global average and max pooling, $\mathbf{d}_c=[\operatorname{GAP}(\Hh_c),\operatorname{GMP}(\Hh_c)]$, where $\Hh_c$ is the feature map of channel $c$. These descriptors are then treated as tokens, and multi-head self-attention is performed across the $C$ channel tokens to explicitly capture inter-channel dependencies. The output is a sigmoid gate $\mathbf{g}$ that reweights the feature map as $\Hh\odot\mathbf{g}$, while this gating is derived from explicit cross-channel attention, and its computational cost does not depend on spatial resolution.

\begin{figure}[ht]\centering
\includegraphics[width=0.95\textwidth]{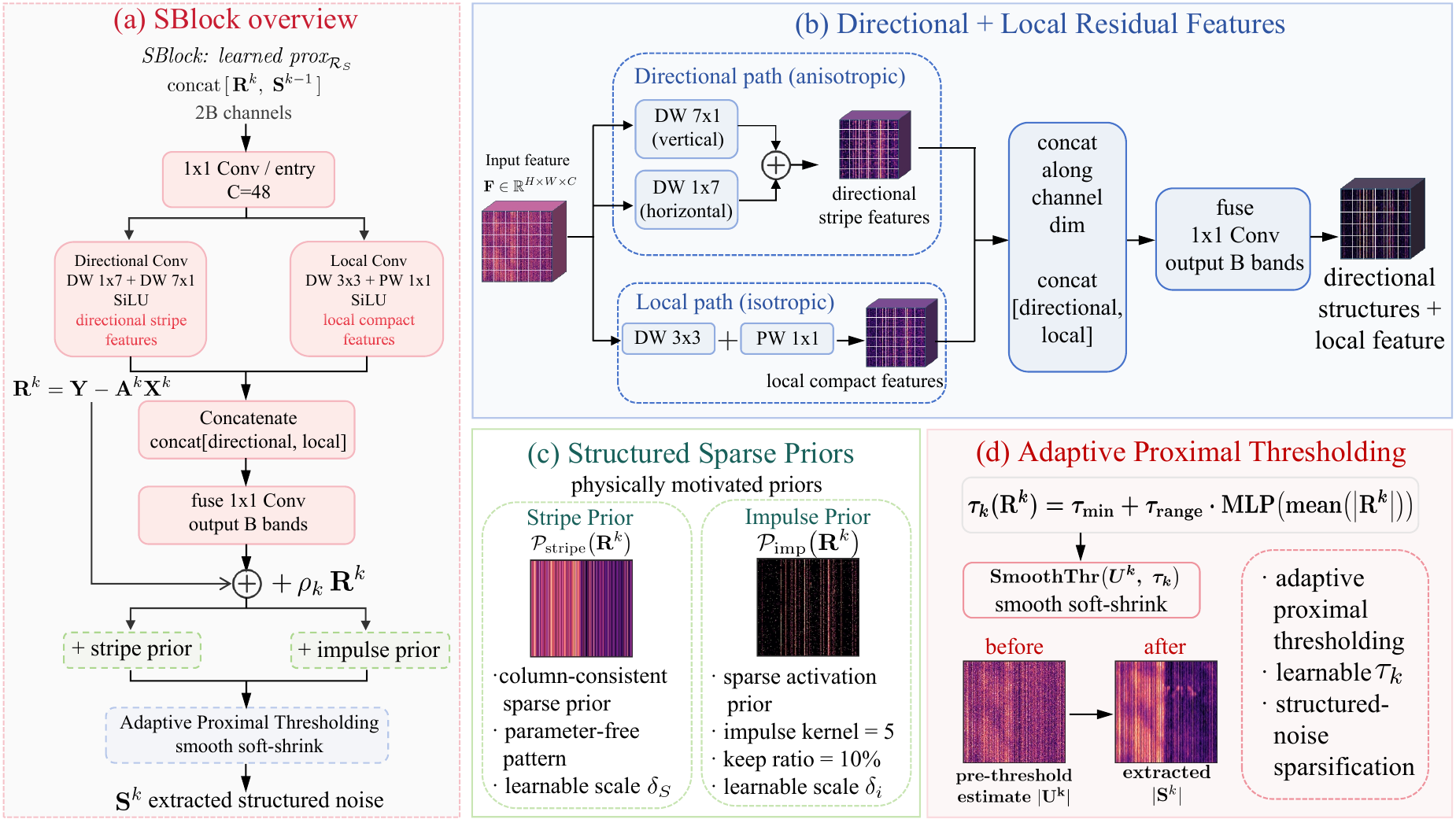}
\caption{SBlock for the S-update. Directional and local branches process
$\R^k=\Y-\A^k\X^k$, combine with column-consistent and sparse priors, and
apply adaptive proximal thresholding to extract $\Snoise^k$.}
\label{fig:sblock}
\end{figure}

\subsection{Structured-noise Update: Structure-Aware Residual Block}% (SBlock)}
\label{sec:sblock}
The structured-noise update applies $\Snoise^{k}=\prox_{\lambda_S\mathcal{R}_S}(\R^k)$ to the reconstruction residual $\R^k=\Y-\A^k\X^k$. %conditioned on the previous estimate $\Snoise^{k-1}$. 
SBlock (Fig.~\ref{fig:sblock}) first maps $\operatorname{Concat}(\R^k,\Snoise^{k-1})$ to width $C$ and extracts two types of residual features. A directional branch uses depthwise $1\times7$ and $7\times1$ convolutions to capture axis-aligned stripe and dead-line structures, while a local branch uses a $3\times3$ convolution to preserve local texture. The branches are fused and added to
a learnable residual skip of $\R^k$:
\begin{equation}
  \widetilde{\Snoise}^{k}
  = \operatorname{Fuse}
    \!\left(\operatorname{Dir}(\cdot),\operatorname{Loc}(\cdot)\right)
    + \rho_k\,\R^k ,
  \label{eq:sraw}
\end{equation}
where $\rho_k$ is initialized to $0.05$.

On top of the learned residual features, we employ two parameter-free structured priors $\mathcal{P}_{\mathrm{stripe}}(\R)$ and $\mathcal{P}_{\mathrm{imp}}(\R)$ to remove the stripe/dead-line noise and impulse noise, respectively. The Column-consistency prior and the Median-residual prior, motivated by~\cite{destripe} and~\cite{medfilt}, respectively, are given by
\begin{align*}
  \mathcal{P}_{\mathrm{stripe}}(\R)
  &= \operatorname{Expand}\!\left(
      \operatorname{Mean}_{h}(\R)
      - \operatorname{Mean}_{w}\!\left(\operatorname{Mean}_{h}(\R)\right)
    \right),\\
  \mathcal{P}_{\mathrm{imp}}(\R)
  &= \operatorname{TopK}_{\kappa}
    \!\left(\R-\operatorname{Median}_{5\times5}(\R)\right)
\end{align*}
where $\operatorname{Mean}_{h}$ averages along the vertical dimension,
$\operatorname{Mean}_{w}$ removes the band-wise global column offset, and the keep ratio $\kappa=0.10$. The learned estimate and the two priors are
combined through learnable non-negative scales $\delta_s$ and $\delta_i$, i.e.,
\begin{equation}
  \mathbf{U}^{k}
  = \widetilde{\Snoise}^{k}
    + \delta_s\,\mathcal{P}_{\mathrm{stripe}}(\R^k)
    + \delta_i\,\mathcal{P}_{\mathrm{imp}}(\R^k).
  \label{eq:sprior_comb}
\end{equation}

%  For stripe and dead-line noise, which is approximately
% consistent along image columns, we use a column-consistency prior~\cite{destripe}
% \begin{equation}
%   \mathcal{P}_{\mathrm{stripe}}(\R)
%   = \operatorname{Expand}
%     \!\left(
%       \operatorname{Mean}_{h}(\R)
%       - \operatorname{Mean}_{w}\!\left(\operatorname{Mean}_{h}(\R)\right)
%     \right),
%   \label{eq:stripe_prior}
% \end{equation}
% where $\operatorname{Mean}_{h}$ averages along the vertical dimension,
% $\operatorname{Mean}_{w}$ removes the band-wise global column offset, and
% $\operatorname{Expand}$ broadcasts the column profile back to
% $B\times H\times W$. For impulse noise,
% which is sparse and locally isolated, we use a median-residual prior~\cite{medfilt}
% \begin{equation}
%   \mathcal{P}_{\mathrm{imp}}(\R)
%   = \operatorname{TopK}_{\kappa}
%     \!\left(\R-\operatorname{Median}_{5\times5}(\R)\right),
%   \label{eq:imp_prior}
% \end{equation}
% with keep ratio $\kappa=0.10$. The two prior maps are parameter-free; only
% their fusion scales are learned. 

Finally, SBlock applies an adaptive proximal-style threshold to sparsify the
structured residual:
\begin{equation}
  \Snoise^{k}
  = \operatorname{SmoothThr}\!\left(\mathbf{U}^{k};\,\tau_k(\R^k)\right),
  \qquad
  \tau_k(\R^k)
  = \tau_{\min}
    + \tau_{\mathrm{range}}\cdot
      \operatorname{MLP}\!\left(\overline{|\R^k|}\right),
  \label{eq:sblock}
\end{equation}
where $\overline{|\R^k|}$ is the band-wise spatial average of $|\R^k|$ and
$\tau_k\in[0.005,0.040]$. The smooth-thresholding operator acts as a
data-adaptive soft-shrinkage step~\cite{ista}, encouraging $\Snoise^k$ to contain sparse
structured corruptions rather than clean image content.

\subsection{Training Loss}
\label{sec:loss}

For clean HSI $\mathbf{C}^{\mathrm{gt}}$ and synthetic structured
noise $\Snoise^{\mathrm{gt}}$, we train AXS-Net with deep supervision over all
unrolled stages:
\begin{equation*}
\mathcal{L}_{\mathrm{total}}
= \sum_{k=1}^{K}\frac{k}{K}
  \!\left(
    w_{c}\,\mathcal{L}^{k}_{\mathrm{clean}}
    + w_{m}\,\mathcal{L}^{k}_{\mathrm{sam}}
    + w_{s}\,\mathcal{L}^{k}_{S}
  \right)
  + w_{o}\,\mathcal{L}^{K}_{\mathrm{cons}} ,
\label{eq:loss}
\end{equation*}
where stage weight $k/K$ places stronger emphasis on later stages, and 
stage-wise losses are defined as
\begin{equation*}
\resizebox{0.95\textwidth}{!}{$\displaystyle
\begin{aligned}
\mathcal{L}^{k}_{\mathrm{clean}}
&= \left\|\A^{k}\X^{k}-\mathbf{C}^{\mathrm{gt}}\right\|_1,
&\qquad
\mathcal{L}^{k}_{S}
&= \left\|\Snoise^{k}-\Snoise^{\mathrm{gt}}\right\|_1,\\
\mathcal{L}^{k}_{\mathrm{sam}}
&= \frac{1}{n}\sum_{i=1}^{n}\arccos
   \frac{\left\langle(\A^{k}\X^{k})_{i},
                    \mathbf{C}^{\mathrm{gt}}_{i}\right\rangle}
        {\left\|(\A^{k}\X^{k})_{i}\right\|_2
         \left\|\mathbf{C}^{\mathrm{gt}}_{i}\right\|_2},
&
\mathcal{L}^{K}_{\mathrm{cons}}
&= \left\|\Y-\A^{K}\X^{K}-\Snoise^{K}
   -\Nnoise^{\mathrm{gt}}\right\|_F^2 .
\end{aligned}
$}
\label{eq:loss_terms}
\end{equation*}
Here $\mathcal{L}_{\mathrm{clean}}$ supervises the reconstructed clean image in
intensity space, and $\mathcal{L}_{\mathrm{sam}}$ preserves per-pixel spectral
fidelity through the spectral angle~\cite{sam}. The term $\mathcal{L}_{S}$ directly
supervises the structured-noise branch, while $\mathcal{L}_{\mathrm{cons}}$
encourages the final decomposition to match the observation model in
Eq.~\eqref{eq:model}. Under the synthetic-noise protocol,
$\Snoise^{\mathrm{gt}}$ and $\Nnoise^{\mathrm{gt}}$ are available from the
noise-generation process. When ground-truth noise components are unavailable,
we omit $\mathcal{L}_{S}$ and $\mathcal{L}_{\mathrm{cons}}$ and train only with
the clean-image supervision. Unless otherwise stated, we set
$w_{c}{=}1.3$, $w_{m}{=}0.2$, $w_{s}{=}0.1$, and $w_{o}{=}0.1$.

% Thus, the reconstruction losses supervise the clean estimate,
% while $\mathcal{L}_{S}$ and $\mathcal{L}_{\mathrm{cons}}$ enforce consistency
% of the structured-noise branch with Eq.~\eqref{eq:model}.

\section{Experiments}
\label{sec:experiments}

\subsection{Experimental Setup}
\label{sec:setup}

\paragraph{Datasets and noise.}
We train our models on ICVL~\cite{icvl} and evaluate on ICVL (50 test scenes),
CAVE~\cite{cave} (32 scenes), and Harvard~\cite{harvard} (26 scenes), where
CAVE and Harvard are used as zero-shot transfer benchmarks. All hyperspectral
tensors consist of 31 bands. We construct five noise settings: (c1) Gaussian,
(c2) Gaussian+stripe, (c3) Gaussian+dead-line, (c4) Gaussian+impulse, and
(c5) a mixture of all four, and use the same fixed noisy cubes for all methods.

\paragraph{Compared methods.}
We conduct a quantitative comparison using MPSNR, MSSIM \cite{ssim}, SAM \cite{sam}, ERGAS~\cite{ergas}, and RMSE for several recent learning-based HSI denoisers under a unified inference setting: T3SC~\cite{t3sc}, MAC-Net~\cite{macnet}, HSDT~\cite{hsdt}, and SSUMamba~\cite{ssumamba}.
Classical model-based methods (BM4D~\cite{bm4d}, LRMR~\cite{lrmr}, NGMeet~\cite{ngmeet}) target different noise assumptions and are not directly comparable under this unified protocol, so we concentrate the comparison on recent learned denoisers. 
All methods are evaluated on the same set of pre-generated noisy cubes (seed $2026$), using identical normalization procedures and metrics. For the learned baselines, we rely on the official model weights and their native inference pipelines, without any additional retuning or test-time augmentation. SSUMamba results are taken from archived predictions on the same
fixed cubes, generated by the officially released SSUMamba (SSCS) checkpoint without any
post-processing or test-time augmentation; re-evaluation in our environment is
blocked by an API incompatibility between BiMamba and \texttt{mamba\_ssm}.

\paragraph{Implementation.}
The architecture is configured as described in Secs.~\ref{sec:framework}--\ref{sec:sblock}, and the training objective follows Sec.~\ref{sec:loss}. We choose a rank of $r{=}8$, a hidden width of $C{=}48$, and $4$ attention heads. Training is carried out for $600$ epochs using Adam (learning rate $3\times10^{-5}$, decreased to $3\times10^{-6}$ during the A-step) on $64\times64$ patches, with a batch size of $4$ and random seed $2026$. We employ weight EMA with a decay factor of $0.9999$ and select the final model according to the best validation MPSNR, using a single NVIDIA A100 GPU. For sliding-window inference, we adopt a $64\times64$ window with a stride of $24$.

\subsection{Comparison with State-of-the-Art}
\label{sec:sota}
\begin{figure}[t]\centering
\includegraphics[width=\textwidth]{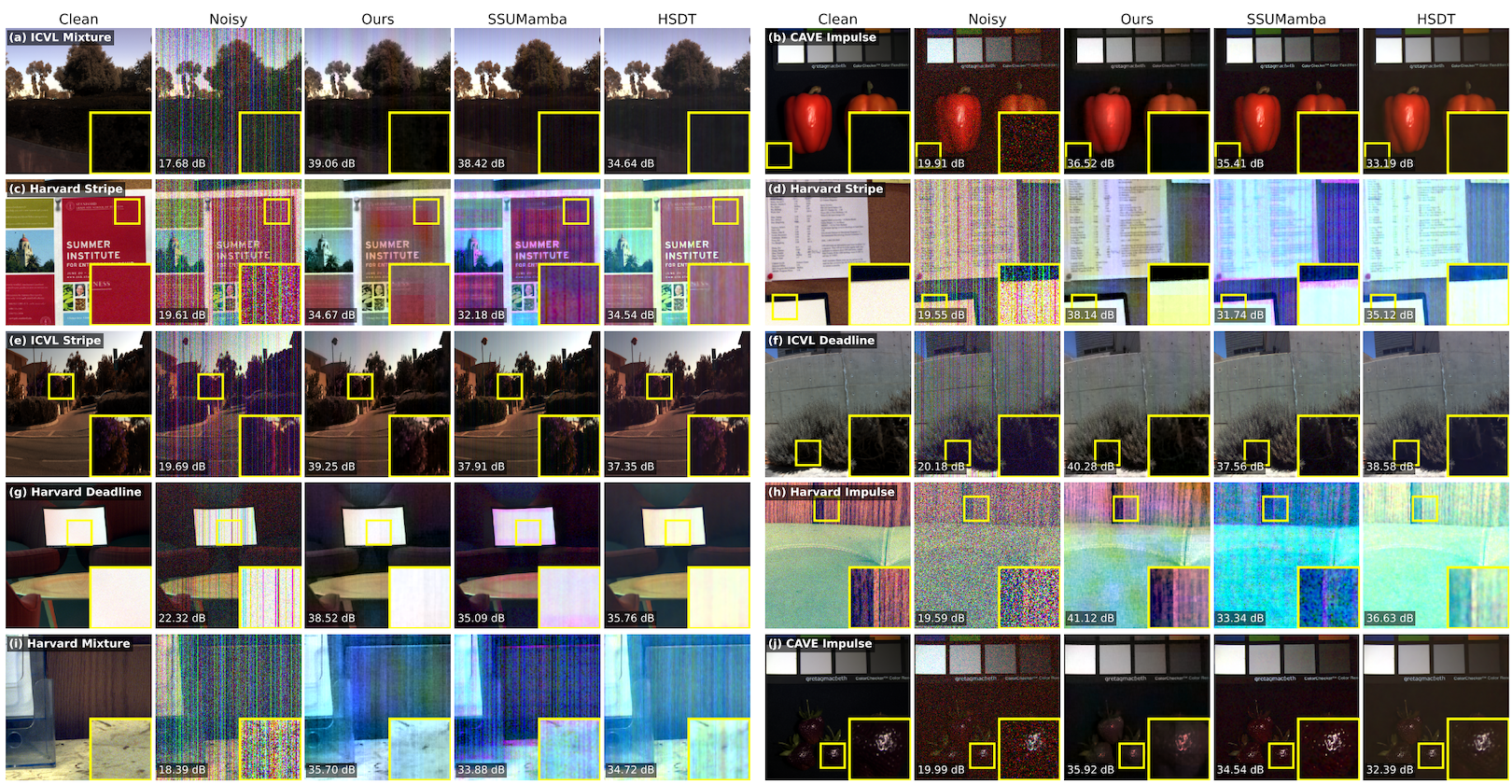}
\caption{Comparisons on ten ICVL, CAVE, and Harvard
scenes for structured noise; CAVE and Harvard are \emph{zero-shot}. Each
block shows Clean/Noisy/Ours/SSUMamba /HSDT with MPSNR (dB).
Panels: (a,e,f) ICVL mixture/stripe/dead-line; (b,j) CAVE impulse; and
(c,d,g,h,i) Harvard stripe/stripe/dead-line/impulse/mixture.}
\label{fig:qual}
\end{figure}

\begin{figure}[t]\centering
\includegraphics[width=\textwidth]{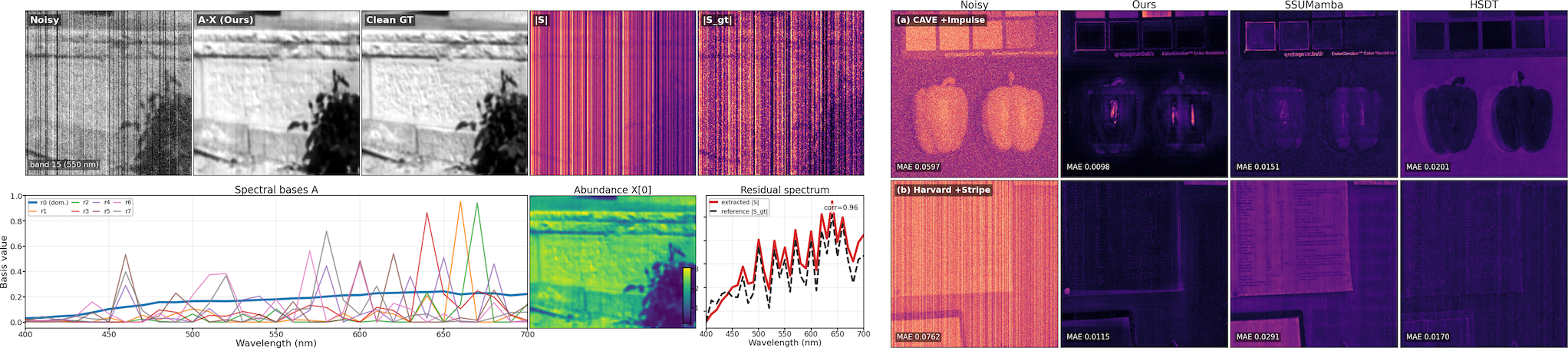}
\caption{\emph{Left:} the results of AXS decomposition including
$\A\X$, $|\Snoise|$, $\A$, and an abundance map. \emph{Right:} zero-shot
band-averaged error maps on CAVE impulse and Harvard stripe noise; darker is
better and values report MAE.}
\label{fig:decomp}
\end{figure}

On the ICVL dataset (Table~\ref{tab:icvl}), the proposed AXS-Net achieves the best performance on all five metrics under every noise configuration and improves MPSNR over SSUMamba by $1.94$/$3.56$/$3.05$\,dB for stripe/dead-line/impulse noise, respectively. In zero-shot transfer (Table~\ref{tab:cross}), it ranks first on all Harvard metrics and all CAVE MSSIM scores, while SSUMamba still achieves a higher MPSNR in CAVE stripe and mixture noise. Figs.~\ref{fig:qual} and~\ref{fig:decomp} illustrate that AXS-Net produces fewer structured residuals and smaller errors in the shown in-domain and zero-shot scenes.
Paired per-scene significance tests (Wilcoxon signed-rank over
$n{=}50/32/26$ scenes for ICVL/CAVE/Harvard, Benjamini--Hochberg FDR at $\alpha{=}0.05$; five noise
settings ${\times}$ four baselines ${\times}$ six metrics---the five reported in
Tables~\ref{tab:icvl}--\ref{tab:cross} plus the correlation coefficient---$={}$
$120$ tests per dataset) give $107$ wins / $8$ ties / $5$ losses on ICVL and $106/9/5$ on
Harvard, whereas the record on CAVE is more balanced at $62/29/29$, consistent
with the weaker CAVE MPSNR transfer noted above. Ties denote differences that
are not significant after FDR correction; because the tables report means while
these tests are paired per scene, a better mean does not always translate into a
significant win.

Our AXS-Net contains only $1.80$\,M parameters, which is substantially fewer than SSUMamba’s $19.35$\,M, though still more than those of the lightweight HSDT, T3SC, and MAC-Net models ($0.52$/$0.91$/$0.43$\,M). When processing a $1392\times1300\times31$ tensor, it requires $14.42$\,s, which is slower than SSUMamba ($7.00$\,s), HSDT ($3.59$\,s), and T3SC ($2.73$\,s), yet faster than MAC-Net ($32.00$\,s), highlighting the computational overhead introduced by its six unrolled stages. These are deployment-time figures under each method's native inference configuration rather than a hardware-normalized ranking of algorithmic complexity.

% ===== Table 1: ICVL full 5-metric (MAIN) =====
\begin{table}[t]\centering\small\setlength{\tabcolsep}{4pt}
\caption{ICVL results ($50$ scenes). Each cell:
MPSNR$\uparrow$/MSSIM$\uparrow$/SAM$\downarrow$ (top),
ERGAS$\downarrow$/RMSE$\downarrow$ (bottom); \textbf{bold}=best.}
\label{tab:icvl}
\vskip-0.25cm
\resizebox{\textwidth}{!}{%
\begin{tabular}{lccccc}
\toprule[1.5pt]
Noise & AXS-Net (Ours) & SSUMamba & HSDT & T3SC & MAC-Net \\
\midrule[0.8pt]
Gaussian & \makecell{\textbf{41.69/0.967/5.61}\\\textbf{14.32/0.0094}} & \makecell{37.63/0.899/14.51\\34.24/0.0141} & \makecell{36.17/0.831/10.73\\93.90/0.0199} & \makecell{33.95/0.805/12.82\\115.87/0.0258} & \makecell{34.56/0.803/13.22\\113.10/0.0257} \\
+Stripe & \makecell{\textbf{39.65/0.957/6.91}\\\textbf{19.67/0.0115}} & \makecell{37.71/0.899/13.18\\36.45/0.0136} & \makecell{36.81/0.842/10.02\\83.17/0.0171} & \makecell{33.71/0.798/13.77\\134.79/0.0261} & \makecell{29.79/0.711/13.76\\150.25/0.0357} \\
+Dead-line & \makecell{\textbf{40.60/0.964/5.92}\\\textbf{15.64/0.0107}} & \makecell{37.04/0.902/14.09\\32.50/0.0155} & \makecell{36.17/0.830/11.02\\96.08/0.0199} & \makecell{33.58/0.801/13.49\\118.90/0.0262} & \makecell{32.54/0.788/14.23\\117.17/0.0285} \\
+Impulse & \makecell{\textbf{41.38/0.964/5.85}\\\textbf{14.75/0.0097}} & \makecell{38.33/0.913/10.17\\32.15/0.0133} & \makecell{35.74/0.826/10.40\\96.72/0.0205} & \makecell{32.50/0.770/12.72\\130.01/0.0290} & \makecell{32.14/0.760/12.89\\130.24/0.0301} \\
Mixture & \makecell{\textbf{38.57/0.947/6.97}\\\textbf{21.56/0.0131}} & \makecell{38.09/0.911/10.84\\33.63/0.0134} & \makecell{36.64/0.845/9.61\\76.43/0.0168} & \makecell{32.25/0.767/13.81\\138.80/0.0285} & \makecell{28.91/0.688/13.84\\156.41/0.0386} \\
\bottomrule[1.5pt]
\end{tabular}}
\end{table}

% ===== Table 2: cross-dataset zero-shot =====
\begin{table}[t]\centering\small\setlength{\tabcolsep}{3pt}
\caption{Zero-shot cross-dataset results. ICVL-trained models are evaluated
on CAVE ($32$) and Harvard ($26$). Each cell:
MPSNR$\uparrow$/MSSIM$\uparrow$/SAM$\downarrow$; \textbf{bold}=best.}
\label{tab:cross}
\vskip-0.25cm
\resizebox{\textwidth}{!}{%
\begin{tabular}{llccccc}
\toprule[1.5pt]
Data & Noise & AXS-Net (Ours) & SSUMamba & HSDT & T3SC & MAC-Net \\
\midrule[0.8pt]
\multirow{6}{*}{CAVE} & Gaussian & \textbf{34.30}/\textbf{0.909}/\textbf{16.55} & 33.69/0.842/29.22 & 33.27/0.745/19.47 & 30.30/0.706/21.23 & 32.24/0.718/22.65 \\
 & +Stripe & 32.40/\textbf{0.867}/24.20 & \textbf{33.76}/0.828/27.15 & 33.46/0.752/\textbf{19.91} & 30.29/0.703/20.57 & 28.34/0.624/23.56 \\
 & +Dead-line & \textbf{33.51}/\textbf{0.901}/\textbf{18.04} & 33.50/0.845/29.04 & 33.22/0.745/19.61 & 29.98/0.702/21.59 & 30.41/0.703/23.45 \\
 & +Impulse & \textbf{34.06}/\textbf{0.905}/\textbf{16.73} & 33.91/0.832/27.05 & 32.97/0.739/19.51 & 29.55/0.676/21.07 & 30.18/0.671/22.85 \\
 & Mixture & 31.80/\textbf{0.854}/25.32 & \textbf{33.66}/0.821/26.29 & 33.09/0.746/\textbf{20.42} & 29.18/0.667/21.01 & 27.26/0.594/24.14 \\
 & \textit{Avg} & 33.21/\textbf{0.887}/20.17 & \textbf{33.70}/0.834/27.75 & 33.20/0.745/\textbf{19.78} & 29.86/0.691/21.09 & 29.69/0.662/23.33 \\
\midrule[0.8pt]
\multirow{6}{*}{Harvard} & Gaussian & \textbf{41.40}/\textbf{0.943}/\textbf{5.74} & 35.53/0.881/14.96 & 35.24/0.824/10.30 & 32.48/0.790/14.60 & 34.51/0.791/13.52 \\
 & +Stripe & \textbf{37.57}/\textbf{0.912}/\textbf{9.77} & 34.96/0.870/14.37 & 35.14/0.827/10.83 & 31.79/0.773/14.57 & 29.54/0.676/15.40 \\
 & +Dead-line & \textbf{40.19}/\textbf{0.936}/\textbf{6.40} & 35.60/0.881/15.11 & 35.25/0.824/10.52 & 32.60/0.792/14.77 & 33.68/0.784/14.23 \\
 & +Impulse & \textbf{41.23}/\textbf{0.942}/\textbf{5.76} & 35.90/0.893/13.08 & 34.95/0.818/10.47 & 31.09/0.746/14.97 & 32.36/0.747/14.16 \\
 & Mixture & \textbf{36.11}/\textbf{0.896}/\textbf{11.09} & 35.28/0.880/13.32 & 34.91/0.822/11.26 & 30.85/0.739/15.19 & 28.96/0.660/16.24 \\
 & \textit{Avg} & \textbf{39.30}/\textbf{0.926}/\textbf{7.75} & 35.45/0.881/14.17 & 35.10/0.823/10.68 & 31.76/0.768/14.82 & 31.81/0.732/14.71 \\
\bottomrule[1.5pt]
\end{tabular}}
\end{table}

\begin{figure}[t]\centering
\includegraphics[width=\textwidth]{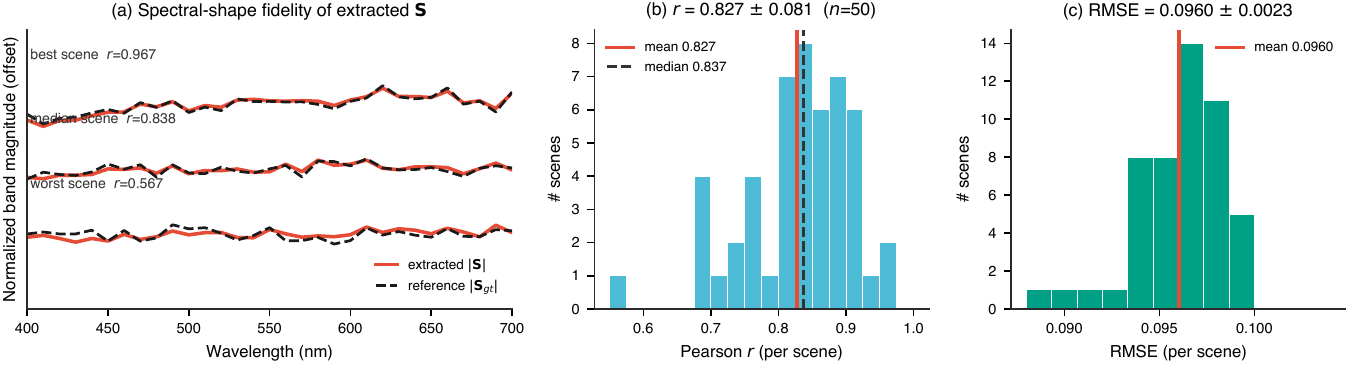}
\caption{Spectral-shape fidelity of the extracted structured noise
$\Snoise$ on ICVL ($n{=}50$): \emph{(a)} extracted versus reference band-magnitude
profiles for the best, median, and worst scenes; \emph{(b)} per-scene Pearson
correlation; \emph{(c)} per-scene RMSE.}
\label{fig:interp}
\end{figure}

\begin{figure}[t]\centering
\includegraphics[width=\textwidth]{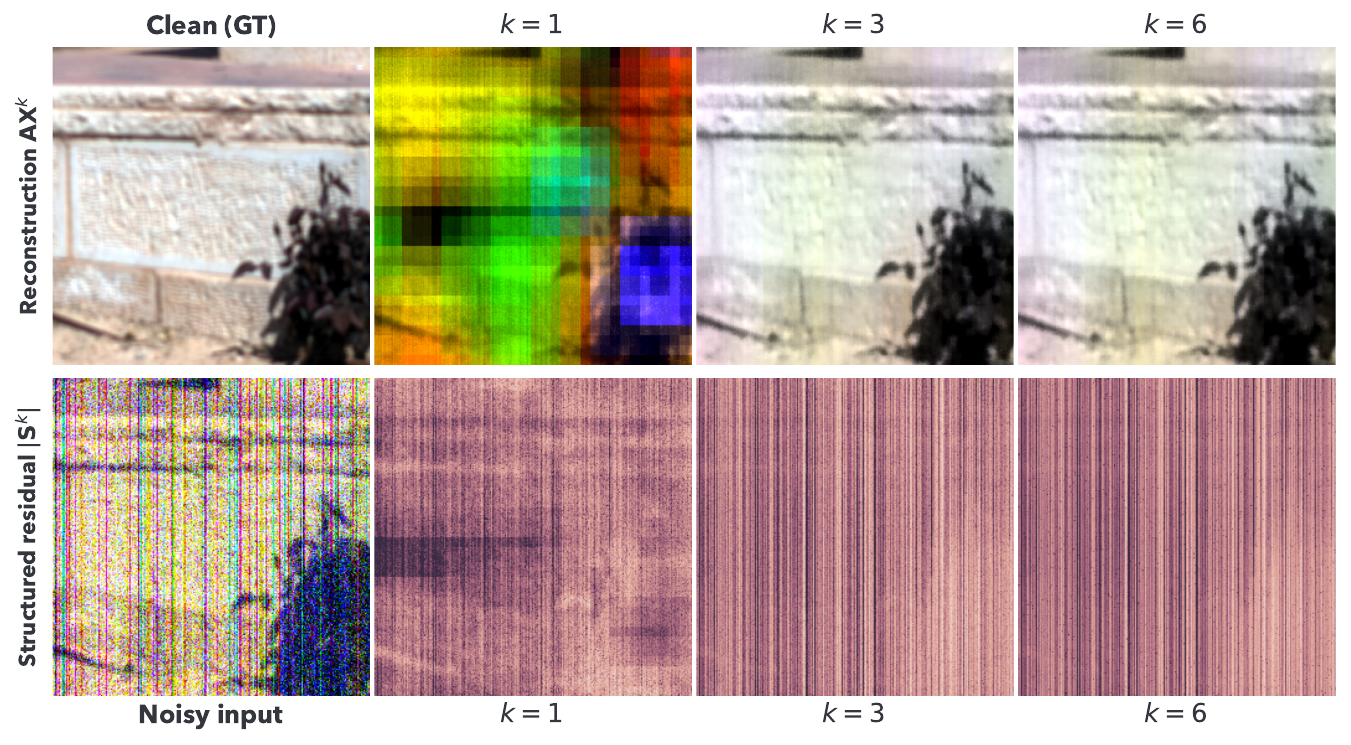}
\caption{{Stage-wise evolution of the decomposition on an ICVL mixture scene.
\emph{Top:} reconstruction $\A^{k}\X^{k}$ at $k{=}1,3,6$ against the clean
reference. \emph{Bottom:} structured residual $|\Snoise^{k}|$ against the noisy
input. The reconstruction sharpens while the residual concentrates onto the
column-consistent stripe pattern.}}
\label{fig:stages}
\end{figure}

\subsection{Interpretability and Structured-Noise Recovery}
\label{sec:interpret}
Fig.~\ref{fig:decomp} (left) shows the decompositions $\A\X$, $|\Snoise|$, $\A$, and a representative abundance map. The recovered bases are smooth and
band-ordered, i.e., they vary gradually with wavelength, and the corresponding
abundance maps preserve clear spatial structure, indicating that the subspace is spectrally organized rather than
composed of arbitrary latent channels. Fig.~\ref{fig:stages} traces this across
the unrolled stages: the reconstruction $\A^{k}\X^{k}$ sharpens from a coarse
piecewise-constant estimate at $k{=}1$ to a faithful image at $k{=}6$, while the
structured residual $|\Snoise^{k}|$ concentrates onto the column-consistent
stripe pattern. We do not claim that the recovered endmembers correspond to identifiable physical materials.

Across the 50 ICVL mixture scenes (Fig.~\ref{fig:interp}), the estimated noise profiles closely follow the reference curves, with a median Pearson correlation of $0.84$ (mean $0.83\pm0.08$; $37/50>0.80$) and an RMSE of $0.096\pm0.002$. Since $\Snoise$ is trained with synthetic supervision, these metrics quantify the consistency of the decomposition under our experimental setting.

\subsection{Ablation Study}\label{sec:ablation}
\paragraph{SSX-Block.}
Table~\ref{tab:ablation} shows the noise-specific effects of each component: spatial attention is most beneficial for Gaussian and mixture noise, while omitting any component leads to a marked drop in performance on impulse noise.

\paragraph{Decomposition components.}
Table~\ref{tab:coreabl} ablates the three structural ingredients of the
unfolding on ICVL mixture noise (c5); each variant is retrained from scratch
with exactly one factor changed. Removing the structured-noise branch
($\Snoise^{k}{=}0$ at every stage) costs $8.15$\,dB MPSNR, freezing the basis at
its PCA initialization costs $10.02$\,dB, and halving the depth from $K{=}6$ to
$K{=}3$ costs $5.22$\,dB---all $8$--$15\times$ larger than the largest single
SSX-Block sub-component effect on the same setting ($0.65$\,dB,
Table~\ref{tab:ablation}). The cost of the $\Snoise$ branch also tracks the
amount of structured content: $0.12$\,dB on pure Gaussian noise
($41.69\!\rightarrow\!41.57$), $2.38$\,dB on stripe noise
($39.65\!\rightarrow\!37.27$), and $8.15$\,dB on the mixture
($38.57\!\rightarrow\!30.42$), indicating that the branch performs
structure-specific work rather than merely adding generic capacity.

\paragraph{SBlock.}
In Table~\ref{tab:coreabl}, removing the impulse prior costs
$1.06$\,dB MPSNR, whereas removing the stripe prior or the local branch leaves
MPSNR essentially unchanged ($+0.10$ and $+0.03$\,dB). The impulse prior is
therefore responsible for the sparse-corruption component, while the other two
act on the recovered $\Snoise$ rather than on the restored image, which our
image-space metrics do not capture. We retain the stripe prior because it is
parameter-free, and the local branch because it supplies the isotropic features
that the directional branch alone cannot represent; $\Snoise$ is itself a
reported output of the model rather than an internal quantity.

\paragraph{Structured-noise supervision ($\mathcal{L}_S$).}
Removing $\mathcal{L}_S$ improves mixture-noise MPSNR by
$0.70$\,dB ($38.57\!\rightarrow\!39.27$) and SAM by $0.92$
($6.97\!\rightarrow\!6.05$; lower is better), indicating that $\mathcal{L}_S$
mainly tethers $\Snoise$ to the supervised reference rather than improving
restoration. We nevertheless report the supervised configuration as our main
model by design: that tethering is what makes the recovered $\Snoise$ comparable
to the reference profile in Fig.~\ref{fig:interp}, and without it the model
would remain a competitive denoiser but lose its interpretable structured-noise
output.

\begin{table}[ht]\centering\small\setlength{\tabcolsep}{5pt}
\caption{SSX-Block ablation on ICVL (MPSNR, dB; $\Delta$ vs.\ full).}
\vskip-0.25cm
\label{tab:ablation}
	\setlength{\tabcolsep}{1.mm}{
		\begin{tabular*}{0.98\hsize}{@{}@{\extracolsep{\fill}}
				lcccc@{}}      %}{ccccccc ccccccc}
		\toprule[1.5pt]
Noise & Full SSX & w/o FFN & w/o Spatial & w/o Spectral \\
\midrule
Gaussian (c1) & 41.69 & 41.60\,($-0.09$) & 40.24\,($-1.45$) & 41.70\,($+0.01$) \\
Impulse (c4)  & 41.38 & 38.02\,($-3.36$) & 38.97\,($-2.41$) & 39.40\,($-1.98$) \\
Mixture (c5)  & 38.57 & 38.46\,($-0.11$) & 37.92\,($-0.65$) & 38.57\,(0.00) \\
\bottomrule[1.5pt]
\end{tabular*}}
\end{table}

\begin{table}[ht]\centering\small\setlength{\tabcolsep}{5pt}
\caption{Decomposition, SBlock, and loss ablation on ICVL mixture noise
(c5, $50$ test scenes, no test-time augmentation), scored with the sliding-window protocol of
Table~\ref{tab:icvl}. Each variant is retrained with the full recipe, changing
exactly one factor; $\Delta$ is MPSNR relative to the full $K{=}6$ model. \textbf{Bold}=best value in each metric column.}
\label{tab:coreabl}
\vskip-0.25cm
\begin{tabular*}{0.98\hsize}{@{}@{\extracolsep{\fill}}lcccc@{}}
\toprule[1.5pt]
Variant & MPSNR$\uparrow$ & MSSIM$\uparrow$ & SAM$\downarrow$ & $\Delta$MPSNR \\
\midrule
Full AXS-Net ($K{=}6$) & 38.57 & 0.947 & 6.97 & -- \\
\midrule
w/o $\Snoise$ branch & 30.42 & 0.821 & 24.00 & $-8.15$ \\
fixed $\A$ (no basis update) & 28.55 & 0.811 & 25.91 & $-10.02$ \\
$K{=}3$ stages & 33.35 & 0.857 & 19.83 & $-5.22$ \\
\midrule
w/o impulse prior & 37.51 & 0.939 & 7.56 & $-1.06$ \\
w/o stripe prior & 38.67 & 0.947 & 6.81 & $+0.10$ \\
w/o local branch & 38.60 & 0.947 & 6.98 & $+0.03$ \\
\midrule
w/o $\mathcal{L}_S$ & \textbf{39.27} & \textbf{0.952} & \textbf{6.05} & $+0.70$ \\
\bottomrule[1.5pt]
\end{tabular*}
\end{table}

\section{Conclusion}
\label{sec:conclusion}
AXS-Net unrolls an $\A$--$\X$--$\Snoise$ decomposition with an
analytic basis step and learned abundance and structured-noise operators. It
achieves strong in-domain and zero-shot results while exposing spectrally
structured intermediate variables. Its main limitations are reliance on
synthetic $\Snoise$ supervision, evaluation confined to synthetically corrupted
data, and weaker MPSNR transfer on strongly shifted CAVE stripe and mixture
cases; unsupervised structured-noise estimation and validation on real-noise
HSIs are left for future work.

\begin{credits}
\subsubsection{\ackname}
This work was supported by Hunan Provincial Natural Science
Foundation of China, under the Science and Technology Innovation Program of
Hunan Province (Project No.~2025JJ60883); by the Hunan Provincial College
Students' Entrepreneurship Training Program (Project No.~S202510530147X); and
by the National College Students' Innovation Training Program (Project
No.~202510530057). The authors also thank the High Performance Computing
Platform of Xiangtan University.
\subsubsection{\discintname}
The authors have no competing interests to declare that are relevant to the
content of this article.
\end{credits}

\FloatBarrier
% ---- Bibliography ----
% Camera-ready: the reference list is generated by BibTeX from refs.bib with
% Springer's splncs04 style, so the source archive ships refs.bib + the
% resulting .bbl, as the ICIG camera-ready instructions ask for.
% Build order: pdflatex -> bibtex -> pdflatex -> pdflatex
\bibliographystyle{splncs04}
\bibliography{refs}

@article{lrmr,
  author  = {Zhang, H. and He, W. and Zhang, L. and Shen, H. and Yuan, Q.},
  title   = {Hyperspectral image restoration using low-rank matrix recovery},
  journal = {{IEEE} Trans. Geosci. Remote Sens.},
  volume  = {52},
  number  = {8},
  pages   = {4729--4743},
  year    = {2014},
  doi          = {10.1109/TGRS.2013.2284280}
}

@article{lrtv,
  author       = {He, W. and Zhang, H. and Zhang, L. and Shen, H.},
  title        = {Total-variation-regularized low-rank matrix factorization for hyperspectral image restoration},
  journal      = {{IEEE} Trans. Geosci. Remote Sens.},
  volume       = {54},
  number       = {1},
  pages        = {178--188},
  year         = {2016},
  doi          = {10.1109/TGRS.2015.2452812}
}

@article{bm4d,
  author  = {Maggioni, M. and Katkovnik, V. and Egiazarian, K. and Foi, A.},
  title   = {Nonlocal transform-domain filter for volumetric data denoising and reconstruction},
  journal = {{IEEE} Trans. Image Process.},
  volume  = {22},
  number  = {1},
  pages   = {119--133},
  year    = {2013},
  doi          = {10.1109/TIP.2012.2210725}
}

@inproceedings{ngmeet,
  author    = {He, W. and Yao, Q. and Li, C. and Yokoya, N. and Zhao, Q.},
  title     = {Non-local meets global: {An} integrated paradigm for hyperspectral denoising},
  booktitle = {{CVPR}},
  pages     = {6861--6870},
  year      = {2019},
  doi          = {10.1109/CVPR.2019.00703}
}

@inproceedings{hsdt,
  author    = {Lai, Z. and Yan, C. and Fu, Y.},
  title     = {Hybrid spectral denoising transformer with guided attention},
  booktitle = {{ICCV}},
  pages     = {13019--13029},
  year      = {2023},
  doi          = {10.1109/ICCV51070.2023.01201}
}

@article{ssumamba,
  author  = {Fu, G. and Xiong, F. and Lu, J. and Zhou, J.},
  title   = {{SSUMamba}: spatial--spectral selective state space model for hyperspectral image denoising},
  journal = {{IEEE} Trans. Geosci. Remote Sens.},
  volume  = {62},
  year    = {2024},
  doi          = {10.1109/TGRS.2024.3446812},
  pages        = {1--14}
}

@inproceedings{t3sc,
  author    = {Bodrito, T. and Zouaoui, A. and Chanussot, J. and Mairal, J.},
  title     = {A trainable spectral-spatial sparse coding model for hyperspectral image restoration},
  booktitle = {{NeurIPS}},
  year      = {2021},
  volume       = {34},
  pages        = {5430--5442}
}

@article{macnet,
  author  = {Xiong, F. and Zhou, J. and Zhao, Q. and Lu, J. and Qian, Y.},
  title   = {{MAC-Net}: {Model}-aided nonlocal neural network for hyperspectral image denoising},
  journal = {{IEEE} Trans. Geosci. Remote Sens.},
  volume  = {60},
  pages   = {1--14},
  year    = {2022},
  doi          = {10.1109/TGRS.2021.3131878}
}

@inproceedings{senet,
  author    = {Hu, J. and Shen, L. and Sun, G.},
  title     = {Squeeze-and-excitation networks},
  booktitle = {{CVPR}},
  pages     = {7132--7141},
  year      = {2018},
  doi          = {10.1109/CVPR.2018.00745}
}

@inproceedings{istanet,
  author    = {Zhang, J. and Ghanem, B.},
  title     = {{ISTA-Net}: {Interpretable} optimization-inspired deep network for image compressive sensing},
  booktitle = {{CVPR}},
  year      = {2018},
  doi          = {10.1109/CVPR.2018.00196},
  pages        = {1828--1837}
}

@article{unmix,
  author  = {Bioucas-Dias, J. M. and Plaza, A. and Dobigeon, N. and Parente, M. and Du, Q. and Gader, P. and Chanussot, J.},
  title   = {Hyperspectral unmixing overview: {Geometrical}, statistical, and sparse regression-based approaches},
  journal = {{IEEE} J. Sel. Top. Appl. Earth Obs. Remote Sens.},
  volume  = {5},
  number  = {2},
  pages   = {354--379},
  year    = {2012},
  doi          = {10.1109/JSTARS.2012.2194696}
}

@article{nmf,
  author  = {Lee, D. D. and Seung, H. S.},
  title   = {Learning the parts of objects by non-negative matrix factorization},
  journal = {Nature},
  volume  = {401},
  number  = {6755},
  pages   = {788--791},
  year    = {1999},
  doi          = {10.1038/44565}
}

@article{ista,
  author  = {Daubechies, I. and Defrise, M. and De Mol, C.},
  title   = {An iterative thresholding algorithm for linear inverse problems with a sparsity constraint},
  journal = {Comm. Pure Appl. Math.},
  volume  = {57},
  number  = {11},
  pages   = {1413--1457},
  year    = {2004},
  doi          = {10.1002/cpa.20042}
}

@inproceedings{swin,
  author    = {Liu, Z. and Lin, Y. and Cao, Y. and Hu, H. and Wei, Y. and Zhang, Z. and Lin, S. and Guo, B.},
  title     = {{Swin} transformer: {Hierarchical} vision transformer using shifted windows},
  booktitle = {{ICCV}},
  year      = {2021},
  doi          = {10.1109/ICCV48922.2021.00986},
  pages        = {9992--10002}
}

@article{destripe,
  author  = {Bouali, M. and Ladjal, S.},
  title   = {Toward optimal destriping of {MODIS} data using a unidirectional variational model},
  journal = {{IEEE} Trans. Geosci. Remote Sens.},
  volume  = {49},
  number  = {8},
  pages   = {2924--2935},
  year    = {2011},
  doi          = {10.1109/TGRS.2011.2119399}
}

@article{medfilt,
  author  = {Hwang, H. and Haddad, R. A.},
  title   = {Adaptive median filters: new algorithms and results},
  journal = {{IEEE} Trans. Image Process.},
  volume  = {4},
  number  = {4},
  pages   = {499--502},
  year    = {1995},
  doi          = {10.1109/83.370679}
}

@article{sam,
  author  = {Kruse, F. A. and Lefkoff, A. B. and Boardman, J. W. and Heidebrecht, K. B. and Shapiro, A. T. and Barloon, P. J. and Goetz, A. F. H.},
  title   = {The spectral image processing system ({SIPS})---interactive visualization and analysis of imaging spectrometer data},
  journal = {Remote Sens. Environ.},
  volume  = {44},
  number  = {2--3},
  pages   = {145--163},
  year    = {1993},
  doi          = {10.1016/0034-4257(93)90013-N}
}

@inproceedings{icvl,
  author    = {Arad, B. and Ben-Shahar, O.},
  title     = {Sparse recovery of hyperspectral signal from natural {RGB} images},
  booktitle = {{ECCV}},
  year      = {2016},
  doi          = {10.1007/978-3-319-46478-7_2},
  pages        = {19--34}
}

@article{cave,
  author  = {Yasuma, F. and Mitsunaga, T. and Iso, D. and Nayar, S. K.},
  title   = {Generalized assorted pixel camera: postcapture control of resolution, dynamic range, and spectrum},
  journal = {{IEEE} Trans. Image Process.},
  volume  = {19},
  number  = {9},
  pages   = {2241--2253},
  year    = {2010},
  doi          = {10.1109/TIP.2010.2046811}
}

@inproceedings{harvard,
  author    = {Chakrabarti, A. and Zickler, T.},
  title     = {Statistics of real-world hyperspectral images},
  booktitle = {{CVPR}},
  pages     = {193--200},
  year      = {2011},
  doi          = {10.1109/CVPR.2011.5995660}
}

@article{ssim,
  author  = {Wang, Z. and Bovik, A. C. and Sheikh, H. R. and Simoncelli, E. P.},
  title   = {Image quality assessment: from error visibility to structural similarity},
  journal = {{IEEE} Trans. Image Process.},
  volume  = {13},
  number  = {4},
  pages   = {600--612},
  year    = {2004},
  doi          = {10.1109/TIP.2003.819861}
}

@book{ergas,
  author    = {Wald, L.},
  title     = {Data Fusion: Definitions and Architectures---Fusion of Images of Different Spatial Resolutions},
  publisher = {Presses de l'\'{E}cole des Mines de Paris},
  address   = {Paris},
  year      = {2002}
}

\end{document}